%% file: root.tex
\documentclass[letterpaper, 10 pt, conference]{ieeeconf}  

\IEEEoverridecommandlockouts                              

\usepackage{graphics} 
\usepackage{epsfig} 
\usepackage{mathptmx} 
\usepackage{times} 
\usepackage{amsmath} 
\usepackage{amssymb}  
\usepackage{multirow}
\usepackage{booktabs}
\usepackage{stfloats}
\usepackage{float}
\usepackage{url}
\usepackage{xcolor}
\usepackage{hyperref}

\title{\LARGE \bf
  IMPACT-CYCLE: A Contract-Based Multi-Agent System for Claim-Level Supervisory Correction of Long-Video Semantic Memory
}

\author{Weitong Kong$^{1\dagger}$, Di Wen$^{1\dagger*}$, Kunyu Peng$^{1,2}$, David Schneider$^{1}$, Zeyun Zhong$^{1}$,\\ Alexander Jaus$^{1}$, Zdravko Marinov$^{1}$,  Jiale Wei$^{1}$, Ruiping Liu$^{1}$, Junwei Zheng$^{1,3}$,\\ Yufan Chen$^{1}$, Lei Qi$^{4}$, and Rainer Stiefelhagen$^{1}$%
\thanks{$^{\dagger}$These authors contributed equally to this work.}%
\thanks{$^{*}$Corresponding author: Di Wen ({\tt\small di.wen@kit.edu}).}%
\thanks{$^{1}$The authors are with Karlsruhe Institute of Technology, 76131 Karlsruhe, Germany.}%
\thanks{$^{2}$The author is with INSAIT, Sofia University “St. Kliment Ohridski”, 1784 Sofia, Bulgaria.}%
\thanks{$^{3}$The author is with ETH Zurich, 8092 Zurich, Switzerland.}%
\thanks{$^{4}$The author is with Technical University of Munich, 80333 Munich, Germany.}%
}

\usepackage{amsmath}
\begin{document}

\maketitle
\thispagestyle{empty}
\pagestyle{empty}

\input{sec/abstract}

\input{sec/introduction}

\input{sec/related}
\input{sec/methodology2}
\input{sec/experiment2}
\input{sec/conclusion}



\bibliographystyle{IEEEtran}
\bibliography{main}

\end{document}

%% file: sec/abstract.tex
\begin{abstract}

Correcting errors in long-video understanding is
disproportionately costly: existing multimodal pipelines
produce opaque, end-to-end outputs that expose no
intermediate state for inspection, forcing annotators to
revisit raw video and reconstruct temporal logic from
scratch. The core bottleneck is not generation quality
alone, but the absence of a supervisory interface through
which human effort can be proportional to the scope of
each error. We present IMPACT-CYCLE, a supervisory
multi-agent system that reformulates long-video
understanding as iterative claim-level maintenance of a
shared semantic memory—a structured, versioned state
encoding typed claims, a claim dependency graph, and a
provenance log. Role-specialized agents operating under
explicit authority contracts decompose verification into
local object–relation correctness, cross-temporal
consistency, and global semantic coherence, with
corrections confined to structurally dependent claims.
When automated evidence is insufficient, the system
escalates to human arbitration as the supervisory
authority with final override rights; dependency-closure
re-verification then ensures correction cost remains
proportional to error scope. Experiments on VidOR show
substantially improved downstream reasoning (VQA:
0.71→0.79) and a 4.8× reduction in human arbitration
cost, with workload significantly lower than manual
annotation. Code will be released at
\href{https://github.com/MKong17/IMPACT_CYCLE}{https://github.com/MKong17/IMPACT\_CYCLE}.
\end{abstract}

%% file: sec/introduction.tex
\section{INTRODUCTION}

Recent multimodal foundation models~\cite{openai2023gpt4,
team2023gemini, alayrac2022flamingo} have achieved strong
performance on short-video benchmarks, yet their behavior
on long, temporally complex videos remains difficult to
trust and harder still to correct. Existing video-to-LLM
pipelines~\cite{li2023videochat, wang2024videollm} are
built around end-to-end generation: the model produces a
response, and if that response is wrong, the only recourse
is to re-prompt or re-annotate from scratch. This creates
a fundamental asymmetry between error scope and correction
cost. A single misidentified attribute can cascade into a
wrong temporal relation, a wrong answer, and a wrong
caption, yet remediation requires restarting the entire
pipeline. In long videos, where temporal dependencies span
hundreds of frames and errors accumulate across multiple
reasoning granularities~\cite{li2023videochat,
sun2022lfvila}, this opacity makes refinement closer to
re-annotation than lightweight editing.

The root cause is not generation quality alone. It is the
absence of a supervisory correction interface: a
representation that is structured enough to localize
errors, transparent enough for a human supervisor to
inspect, and editable in a way that confines downstream
re-verification to only the affected components. Prior work
on video scene graphs~\cite{ji2020actiongenome,
yang2023pvsg, wu2025psg4d} and graph-grounded
reasoning~\cite{hudson2019gqa, grunde2021agqa} treats
structured representations as static prediction endpoints.
Once generated, these outputs cannot be selectively
revised: when an error is detected, the entire graph must
be regenerated rather than surgically corrected.
Claim-checking approaches~\cite{thauvin2023sgverify}
detect inconsistencies but provide no mechanism for
targeted repair. Human-in-the-loop annotation
methods~\cite{vondrick2011videoal, qiao2023hvsa,
delatolas2024whatandhow, vujasinovic2024strikethebalance}
reduce labeling cost but address the problem of acquiring
new annotations, not the structural correction of
model-generated semantic state. The gap between these
directions, namely detection without repair, and labeling
without correction, is precisely what motivates this work.

We reformulate long-video understanding as a supervisory
correction problem. Rather than asking how to generate a
more accurate scene graph, we ask how a human supervisor
can efficiently maintain the semantic integrity of a long
video by editing a revisable, claim-level memory, without
re-prompting or re-annotating from scratch. To this end,
we present IMPACT-CYCLE, a supervisory multi-agent system
built around a shared semantic memory that encodes typed
claims, their dependency structure, and a provenance log.
This memory serves as the common substrate for both
automated verification and human oversight, exposing
intermediate reasoning as an inspectable and editable
state rather than an opaque final output.

Verification is carried out by role-specialized agents
operating under explicit authority contracts, covering
local object-relation correctness, cross-temporal
consistency, and global semantic coherence. When automated
evidence is insufficient, the system escalates to human
arbitration, positioning the human not as a fallback but
as the supervisory authority with final override rights.
After each human edit, re-verification is confined to the
dependency closure of the modified claims, ensuring that
correction cost remains proportional to error scope rather
than to video length.

This paper makes three contributions:
\begin{itemize}
    \item We introduce a formally defined supervisory semantic memory with typed claims, dependency structure, and provenance, which supports targeted error localization and localized re-verification beyond static scene graph representations.
    
    \item We propose a contract-based multi-agent supervisory architecture comprising five role-specialized agents, among which three form the multi-view verification stage and two serve as control-layer components for memory construction and arbitration.
    
    \item We cast the human as the supervisory authority and restrict post-edit re-verification to the structural neighborhood of each correction, so that human effort scales with error scope. This design reduces arbitration cost by approximately 4.8$\times$ relative to full re-annotation, as supported by quantitative evaluation and a pilot user study.
\end{itemize}

%% file: sec/related.tex
\section{Related Work}

\subsection{Structured Video Graphs}

Scene graph research has progressively tightened the
coupling between visual perception and structured
language, from static image triplets~\cite{johnson2015image,
krishna2017visualgenome, li2017sgg_regioncaption} to
spatio-temporal video graphs with increasingly precise
grounding~\cite{ji2020actiongenome, yang2022psg,
yang2023pvsg, wu2025psg4d, wu2025usg}. Generation
quality has advanced substantially through temporal
modeling~\cite{cong2021sttran, teng2021trace},
anticipatory pretraining~\cite{li2022apt},
debiasing~\cite{nag2023tempura, peddi2025robustvsgg},
and open-vocabulary formulations~\cite{wang2024oed,
li2024pixels2graphs}. Yet across this entire
progression, the scene graph remains a terminal output:
generated once, consumed downstream, and replaced
wholesale when wrong. We depart from this convention by
treating the graph as a revisable semantic memory that
verification agents can read, challenge, and surgically
correct.

\subsection{Cross-Task Evidence}

Structured representations have been shown to support
compositional reasoning across VQA~\cite{hudson2019gqa,
haurilet2019dyngraph, grunde2021agqa,
mao2022dynamicmultistep}, language-conditioned graph
inference~\cite{hu2019lcgn, shi2019xgqa}, and
captioning~\cite{anderson2016spice, yang2019sgae,
zhong2021nls, lu2025comprecap, chu2025videocapsg},
with each task treating the graph as an independent
read-only input. The closest precedent to our approach
is Thauvin and Herbin~\cite{thauvin2023sgverify}, who
detect graph inconsistencies via per-claim VQA on still
images, but provide no repair mechanism and no temporal
reasoning. We extend this paradigm to video by unifying
single-turn VQA, multi-turn VQA, and captioning as
three structurally constrained evidence views over a
shared writable claim space, enabling not just detection
but targeted correction. Related work on egocentric
understanding~\cite{li2026egocross, fu2025objectrelator}
and multimodal anticipation~\cite{peng2025hopadiff,
zhong2023afft} is complementary.

\subsection{Human-in-the-Loop Refinement}

The problem of making model outputs correctable rather
than merely accurate has been studied across several
communities. In annotation and segmentation, active
learning has addressed cost through sample
selection~\cite{vondrick2011videoal, qiao2023hvsa},
joint frame-and-type optimization~\cite{delatolas2024whatandhow},
and uncertainty-triggered interaction in long-video
settings~\cite{vujasinovic2024strikethebalance};
interactive correction has been systematically
characterized in medical imaging as a paradigm in which
human edits propagate through structured
representations rather than triggering full
reprocessing~\cite{marinov2024deep}. Within scene
graphs specifically, uncertainty has been characterized
at the predicate and relation level via Bayesian,
evidential, and conformal formulations~\cite{yang2021ambiguity,
li2023uncertaintysgg, sun2023edal, nag2025conformal},
and structured access has been shown to admit localized
corrections in preference to full
reannotation~\cite{kunzel2025visualxai,
xie2025localcorrections}. These contributions remain
orthogonal: none links selection, uncertainty
quantification, and repair into a single correction
loop. Our framework unifies all three into a
claim-level arbitration stage that ranks open claims
by utility, solicits targeted human judgment, and
propagates corrections only through the dependency
closure of affected claims, making human supervision
both minimal and structurally complete. Complementary
work spans multi-agent industrial
coordination~\cite{wen2025mica}, assistive
perception~\cite{liu2023opensu}, wearable
deployment~\cite{wen2025snapsegmentdeploy}, open-set
recognition~\cite{peng2024ossar}, and domain-agnostic
activity understanding~\cite{schneider2022posecl}.

%% file: sec/methodology2.tex
\section{Methodology}
\label{sec:method}

\subsection{System Overview}

\begin{figure*}[t]
\centering
\includegraphics[width=0.8\linewidth]{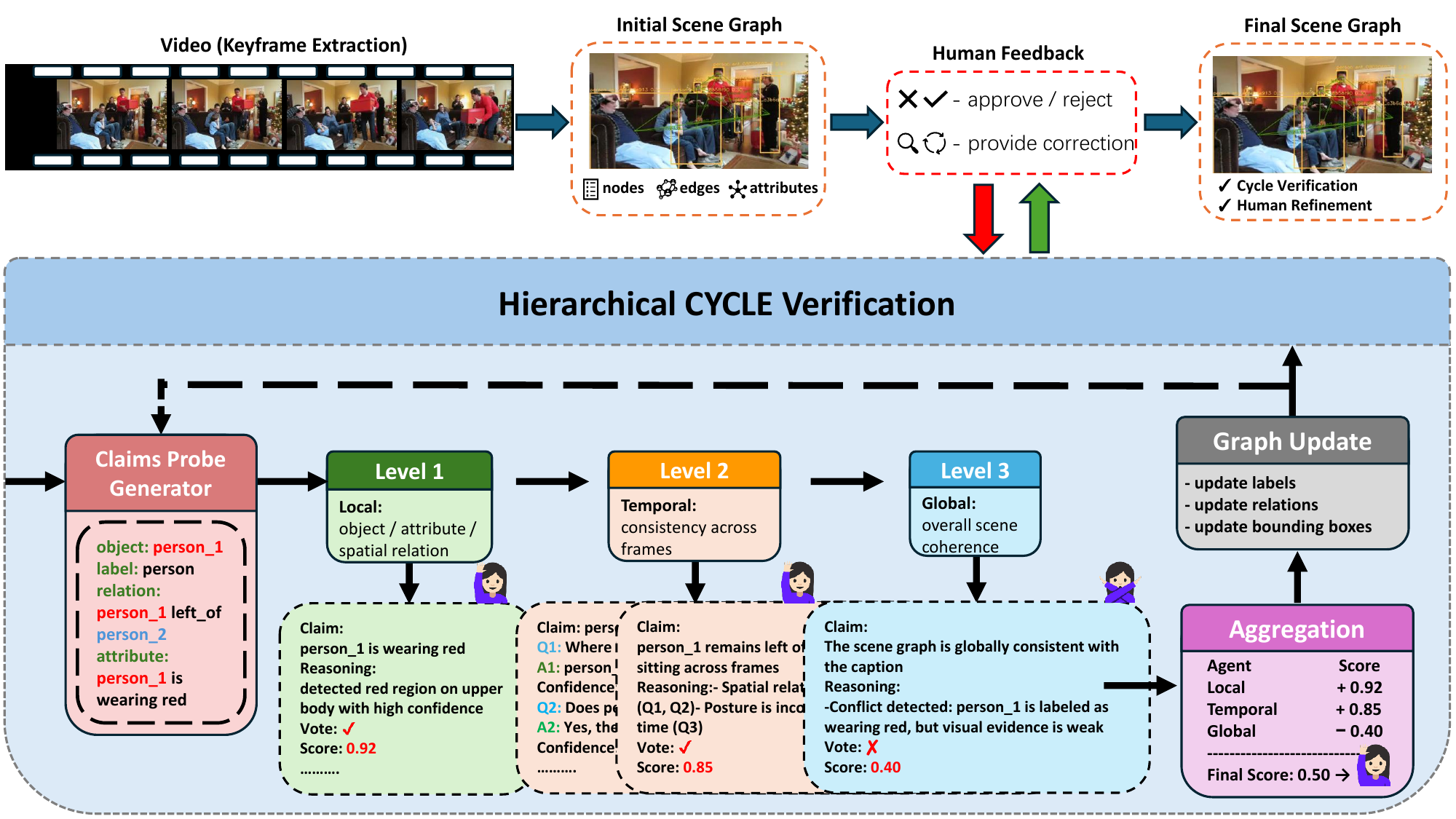}
\caption{%
  Overview of IMPACT-CYCLE. Starting from keyframe-grounded
  scene graphs, the system decomposes graph elements into typed
  atomic claims and instantiates structured verification probes.
  Three verification agents---local grounding, temporal
  consistency, and global semantic audit---provide complementary
  evidence over a shared, revisable semantic memory. Their
  outputs are fused by the Arbitration Agent; unresolved
  conflicts are escalated to the human supervisor, whose edits
  trigger dependency-closure re-verification rather than a
  full pipeline rerun.
}
\label{fig:pipeline}
\end{figure*}

Current video-to-LLM pipelines treat scene understanding as a
one-shot generation problem: a model produces an output, and
if that output contains an error, the only recourse is full
regeneration. IMPACT-CYCLE replaces this paradigm with a
\emph{supervisory correction loop} built around a shared,
versioned semantic memory. The central intuition is that errors
in long-video understanding are rarely total---they are
localized to specific claims about specific entities or
relations---and correction should be equally localized.

\paragraph{Semantic memory}
We define the video-level semantic memory as a four-tuple
\begin{equation}
  \mathcal{M} = (\mathcal{G},\, \mathcal{C},\, \mathcal{D},\, \Pi),
\end{equation}
where $\mathcal{G}$ is the video-level scene graph state,
maintained as a union of frame-level slices $\mathcal{G}_t$
and updated in place throughout verification;
$\mathcal{C}$ is the typed claim set derived from $\mathcal{G}$;
$\mathcal{D}$ is the claim dependency graph encoding structural
dependencies among claims; and $\Pi$ is the provenance log,
which records every agent decision and human edit as a
timestamped, attributed entry. $\Pi$ does not enter the belief
computation directly, but supports auditability, rollback, and
user inspection---properties that are essential for a
supervisory system in which human overrides must be traceable
and reversible.

\paragraph{Agent architecture}
IMPACT-CYCLE comprises five agents operating over $\mathcal{M}$
under explicit authority contracts. The \textbf{Memory
Constructor} initializes $\mathcal{M}$ from raw video with
write authority restricted to initialization. The three
\textbf{verification agents}---Local Grounding, Temporal
Consistency, and Global Semantic Audit---each hold flag-only
authority: they produce structured evidence but cannot
directly modify $\mathcal{G}$. The \textbf{Arbitration Agent}
holds post-fusion write authority: it synthesizes multi-role
evidence, resolves conflicts, and decides whether to accept,
rewrite, or escalate each claim. The Memory Constructor and
Arbitration Agent are control-layer components; only the three
middle agents constitute the multi-view verification stage
discussed in Section~\ref{sec:verification}. The
\textbf{Human Supervisor} holds apex authority, with the
right to override or lock any claim in $\mathcal{M}$; by
default, interaction is restricted to structured queries to
minimize cognitive load, with full override available as an
exceptional supervisory privilege.

\subsection{Memory Construction}

The Memory Constructor initializes $\mathcal{M}$ from a small
set of keyframes $\{I_t\}$ sampled from video $V$.

\paragraph{Keyframe selection}
Candidate frames are derived from motion-driven temporal
segments estimated via scene-change and optical-flow signals,
prior to and independently of any claim decomposition.
For each segment $[t_s, t_e]$, dynamic intervals---those
exhibiting significant motion or state change---contribute up
to three key positions: start, midpoint, and end; stable
intervals contribute the midpoint alone. If the resulting
candidate set exceeds a global budget of $K$ frames, uniform
subsampling is applied. In practice, $K \in \{3,\ldots,5\}$
provides efficient yet representative coverage.

\paragraph{Initial graph construction}
For each keyframe $I_t$, an open-vocabulary proposal pipeline
generates object candidates, filtered by geometric validity and
detection confidence. Retained instances are assigned canonical
labels, spatial extents, and optional attributes; pairwise
relations are inferred to complete the frame-level slice:
\begin{equation}
  \mathcal{G}_t^{(0)} = (\mathcal{N}_t,\, \mathcal{E}_t,\,
  \mathcal{A}_t).
\end{equation}
The video-level graph state $\mathcal{G}^{(0)}$ is initialized
as the union of these frame-level slices, with cross-frame
entity correspondence established by spatial overlap and label
consistency. Separating identity, relations, and attributes
into distinct fields allows each to be revised independently
in later stages.

\paragraph{Claim decomposition}
The initial graph is decomposed into atomic typed claims, each
representing a verifiable hypothesis over a single graph element:
\begin{equation}
  \mathcal{C}_t =
  \mathcal{C}_t^{\mathrm{exist}} \cup
  \mathcal{C}_t^{\mathrm{label}} \cup
  \mathcal{C}_t^{\mathrm{attr}} \cup
  \mathcal{C}_t^{\mathrm{rel}},
\end{equation}
corresponding to entity existence, canonical label assignment,
attribute assertions, and subject--predicate--object relations,
respectively.

\paragraph{Dependency graph}
For each pair of claims $(c, c')$, we add a directed edge
$c \to c'$ in $\mathcal{D}$ if a revision to $c$ structurally
constrains $c'$, via shared entity identity, shared relation
endpoints, or overlapping temporal extents. The dependency graph governs which claims must be re-verified after any accepted edit.

\subsection{Three-Agent Verification over Shared Memory}
\label{sec:verification}

Among the five system agents, the Local Grounding, Temporal
Consistency, and Global Semantic Audit agents constitute the
multi-view verification stage. Each evaluates claims from a
distinct observational scope, producing evidence that is
complementary by construction. Formally, at iteration $r$,
agent $a$ produces:
\begin{equation}
  z_{a,c}^{(r)} = \psi_a\!\left(I_t,\, \mathcal{G}^{(r-1)},\,
  c\right),
\end{equation}
implemented as:
\begin{equation}
  \psi_a\!\left(I_t, \mathcal{G}^{(r-1)}, c\right)
  = \mathrm{Parse}\!\left(
      \mathrm{MLLM}\!\left(P_a\!\left(I_t,
      \mathcal{G}^{(r-1)}, c\right)\right)
    \right),
\end{equation}
where $P_a$ is a role-specific prompt template and
$\mathrm{Parse}(\cdot)$ extracts a structured evidence tuple
$(a, \delta_{a,c}, s_{a,c})$ from the model response.
All three agents use GPT-4V as the shared multimodal backbone,
with details provided in Section~\ref{sec:experiments}.

\paragraph{Local Grounding Agent}
Operating on individual keyframes, this agent evaluates
existence, label validity, attribute correctness, and
frame-local spatial relations through direct single-turn
probes. Its observation scope is the narrowest but its
grounding precision the highest.

\paragraph{Temporal Consistency Agent}
Operating over motion-driven temporal segments and the
corresponding temporal subgraph, this agent evaluates whether
relational and persistent attribute claims hold across sampled
frames. Ambiguous or structurally dependent claims are handled
through multi-turn probing, where each follow-up question is
conditioned on prior answers within the same segment.

\paragraph{Global Semantic Audit Agent}
This agent issues structured queries against both the current
keyframe bundle and a scene-level caption generated by the
MLLM from those keyframes directly---not from the graph
alone---so that its evidence is image-grounded rather than
a textual paraphrase of $\mathcal{G}$. The graph caption
serves as a query scaffold that directs the agent's attention
to specific claim regions; the primary evidence source remains
the visual input. This design allows the agent to detect
globally contradictory or factually unsupported assertions that
neither local nor temporal evidence can surface, while
maintaining independence from the graph state under audit.

\subsection{Role-Aware Evidence Fusion}

Different verification roles have systematically different
reliability profiles across claim types: local grounding is
precise for existence and label claims but cannot assess
cross-frame coherence; temporal reasoning excels at relational
consistency but may miss fine-grained attribute errors;
caption-level audit captures global incoherence but is
imprecise at the label level. Uniform aggregation would
over-trust weak signals and under-use strong ones.

For each claim $c$, the three agents contribute:
\begin{equation}
  \mathcal{Z}(c) = \{(a,\, \delta_{a,c},\, s_{a,c})\},
\end{equation}
where $\delta_{a,c} \in \{-1, 0, +1\}$ denotes contradiction,
abstention, or support, and $s_{a,c} \in [0,1]$ is the
associated confidence. We aggregate into directional scores:
\begin{align}
  S^{+}(c) &= \textstyle\sum_a \omega_a(c)\,s_{a,c}
              \cdot \mathbf{1}[\delta_{a,c}=+1], \\
  S^{-}(c) &= \textstyle\sum_a \omega_a(c)\,s_{a,c}
              \cdot \mathbf{1}[\delta_{a,c}=-1], \\
  S^{0}(c) &= \textstyle\sum_a \omega_a(c)\,s_{a,c}
              \cdot \mathbf{1}[\delta_{a,c}=0].
\end{align}
The role-aware weight $\omega_a(c) = \lambda_{r(a),\tau(c)}$
is defined by a fixed matrix $\lambda$ over role--claim-type
pairs (Table~\ref{tab:role_weights}), calibrated on a
held-out development set disjoint from all evaluation
benchmarks and kept fixed across all experiments. For label claims, the \emph{Global Semantic Audit} agent is assigned zero weight, since caption-level evidence is less reliable than direct visual grounding for object identity.

\begin{table}[t]
\centering
\small
\caption{Role-aware weight matrix $\lambda_{r(a),\tau(c)}$.
  }
\label{tab:role_weights}
\begin{tabular}{lcccc}
\toprule
Role & Exist & Label & Attr & Rel \\
\midrule
Local Grounding        & 1.00 & 1.00 & 0.90 & 1.00 \\
Temporal Consistency   & 0.80 & 0.80 & 0.70 & 0.80 \\
Global Semantic Audit  & 0.70 & 0.00 & 0.60 & 0.70 \\
\bottomrule
\end{tabular}
\end{table}

\subsection{Belief Update and Graph Revision}

\paragraph{Belief update}
The Arbitration Agent computes a revised belief for each
binary-style claim as:
\begin{equation}
  p^{(r)}(c) =
  \frac{\varepsilon + S^{+}(c)}
       {2\varepsilon + S^{+}(c) + S^{-}(c)},
\end{equation}
where $\varepsilon > 0$ is a small stabilizing constant.
The abstention mass $S^{0}(c)$ enters implicitly: it reduces
total directional evidence without biasing the outcome,
widening the region in which the Arbitration Agent defers
rather than commits.

\paragraph{Constrained correction}
For correction claims---label or relation replacement---the
Arbitration Agent selects among ontology-valid candidates
$\mathcal{O}(c)$:
\begin{equation}
  Q(o \mid c)
  = \sum_a \omega_a(c)\,s_{a,c}
    \cdot \mathbf{1}[o_{a,c} = o],
\quad
  \hat{o} = \arg\max_{o \in \mathcal{O}(c)} Q(o \mid c).
\end{equation}
A correction is accepted only if $Q(\hat{o} \mid c)$ exceeds
both a revision threshold $\theta$ and the score of the current
graph value, ensuring that revisions are conservative and
evidence-grounded. All accepted updates are written back to
$\mathcal{G}$ with provenance entries appended to $\Pi$.

\paragraph{Termination}
The refinement loop runs for at most $r_{\max}$ iterations
or until convergence, defined as no claim changing state
between consecutive rounds. In practice $r_{\max} = 2$
suffices; the ablation in Section~\ref{sec:experiments}
confirms that most gains accrue in the first round.

\subsection{Human Arbitration and Localized Re-Verification}

When the Arbitration Agent cannot reach a confident decision,
it escalates the claim to the human supervisor. The default
interaction mode is restricted to structured queries---binary
validation or candidate selection from $\mathcal{O}(c)$---to
minimize cognitive load and ensure judgments are comparable
across sessions. Full supervisory override of any claim
in $\mathcal{M}$ remains available as an exceptional
privilege, representing the apex authority in the system.

Escalation priority is determined by an arbitration utility
\begin{equation}
  u(c) = \alpha \cdot \mathrm{unc}(c)
         + \beta  \cdot \mathrm{conflict}(c)
         + \gamma \cdot \mathrm{impact}(c),
\end{equation}
where $\mathrm{unc}(c) = S^{0}(c)$ measures epistemic
abstention, $\mathrm{conflict}(c) = \min(S^{+}(c), S^{-}(c))$
measures simultaneous support and contradiction pressure
(genuine inter-role disagreement rather than one-sided
rejection), and $\mathrm{impact}(c)$ is the out-degree of
$c$ in $\mathcal{D}$, serving as a simple structural proxy
for downstream correction cost. Claims are escalated when
$u(c) < \theta_u$.

After each accepted edit to a claim set $\mathcal{Q}$,
only the dependency closure
\begin{equation}
  \Gamma(\mathcal{Q}) =
  \mathcal{Q} \cup
  \{c' : c' \sim c,\ c \in \mathcal{Q}\}
\end{equation}
is re-verified, where $c' \sim c$ denotes dependency in
$\mathcal{D}$ via shared entities, relations, or temporal
extent. This is the mechanism by which correction cost is
kept proportional to error scope: a localized edit triggers
a localized rerun, not a full pipeline restart.
Since $r$, $K$, $P_s$, and $P_m$ are all small and fixed,
total model calls scale as
$\mathcal{O}(r \cdot K \cdot (P_s + P_m + 1))$
with respect to the verification budget rather than raw
video length.

%% file: sec/experiment2.tex
\section{Experiments}
\label{sec:experiments}

\paragraph{Setup}
We evaluate IMPACT-CYCLE on VidOR~\cite{shang2019annotating}, a long-video benchmark with temporally grounded relation annotations. Hyperparameters are selected on a held-out development split, yielding $r_{\max}=2$, $K=5$, $P_s \leq 5$, and $P_m \leq 2$. GPT-4V (\texttt{gpt-4-vision-preview}) is used as the shared verification backbone across all roles. In the main experiments, human arbitration is simulated with oracle decisions to provide an upper bound on the benefit of supervisory intervention; real-user behavior is examined separately in the user study. VQA evaluation uses the same MLLM for question generation and answer assessment, which may introduce self-evaluation bias. Captioning and multi-turn VQA are used in our framework as verification evidence views, but are not separately evaluated as downstream tasks in this paper.

\paragraph{Metrics}
We report three groups of metrics. For structural quality, we use \textbf{Entity Accuracy} and \textbf{Graph Edit Distance (GED)}. For verification behavior, we report \textbf{Inv.Probe}, the fraction of probes that fail to produce a parseable structured response; \textbf{Uncert.}, the fraction of claims remaining in the abstention state $\delta = 0$ after all roles have responded; \textbf{ClaimAgr}, the inter-role agreement rate on the same claim; \textbf{Resolve Score}, the fraction of disputed claims resolved without human escalation; and \textbf{Human-Q/F}, the number of arbitration queries issued per frame. For downstream utility, we report \textbf{VQA Accuracy}.

\paragraph{Main Results}

\begin{table}[t]
\centering
\small
\caption{Structural and downstream performance on VidOR.}
\label{tab:main_results}
\begin{tabular}{lccc}
\toprule
Method & Entity Acc $\uparrow$ & GED $\downarrow$
& VQA $\uparrow$ \\
\midrule
MLLM (no graph)           & --    & --    & 0.66 \\
Initial Graph             & 0.900 & 0.182 & 0.71 \\
After Verification (Ours) & \textbf{0.931}
                          & \textbf{0.179}
                          & \textbf{0.79} \\
\bottomrule
\end{tabular}
\end{table}

Table~\ref{tab:main_results} shows modest GED improvement
(0.182$\rightarrow$0.179) but substantial VQA gain
(0.71$\rightarrow$0.79). This asymmetry is expected: GED
is insensitive to relation mismatches and attribute errors
that leave graph topology unchanged but strongly disrupt
downstream inference. The VQA gain is statistically
consistent across videos ($p{<}0.05$, $n{=}240$), and
Resolve Score correlates strongly with VQA accuracy
($r{=}0.72$, video level), confirming that claim-level
dispute resolution reliably predicts downstream reasoning
quality.

\paragraph{Verification Behavior}

\begin{table}[t]
\centering
\small
\setlength{\tabcolsep}{3.5pt}
\renewcommand{\arraystretch}{1.05}
\caption{Verification behavior across backends and density
  regimes.}
\label{tab:structure_verification_all}
\resizebox{\columnwidth}{!}{%
\begin{tabular}{llccccc}
\toprule
Model & Density
& Inv.Probe $\downarrow$ & Uncert. $\downarrow$
& ClaimAgr $\uparrow$ & Resolve $\uparrow$
& Human-Q/F $\downarrow$ \\
\midrule
\multirow{3}{*}{GPT-4V}
& Low    & 0.377 & 0.347 & 0.750 & 0.250 & 0.019 \\
& Med.   & 0.488 & 0.448 & 0.475 & 0.256 & 1.350 \\
& High   & 0.655 & 0.611 & 0.521 & 0.274 & 2.000 \\
\midrule
\multirow{3}{*}{Gemini Pro}
& Low    & 0.414 & 0.363 & 0.400 & 0.600 & 0.005 \\
& Med.   & 0.442 & 0.371 & 0.491 & 0.354 & 0.690 \\
& High   & 0.495 & 0.453 & 0.438 & 0.389 & 2.026 \\
\bottomrule
\end{tabular}}
\end{table}

Density regimes are defined by claim count: low
($|\mathcal{C}|{<}10$), medium ($10{\leq}|\mathcal{C}|{<}20$),
high ($|\mathcal{C}|{\geq}20$).
Table~\ref{tab:structure_verification_all} reveals three
consistent patterns across both backends. First, Inv.Probe,
Uncert., and H-Q/F increase monotonically with density,
confirming that escalation effort scales proportionally with
structural complexity. Second, ClaimAgr decreases with
density as inter-role disagreement grows---disagreement that
the Arbitration Agent routes to human arbitration rather than
resolving arbitrarily. Third, Resolve Score is highest in
low-density regimes where claims are individually simpler;
in high-density graphs, interdependent claims are
deliberately withheld from automatic resolution by the
dependency-closure design. Gemini Pro achieves higher
Resolve Score at low density (0.600 vs.\ 0.250 for GPT-4V)
but converges to similar H-Q/F at high density, suggesting
backend choice has diminishing influence as structural
difficulty increases.

\paragraph{Ablation Study}

\begin{table}[t]
\centering
\small
\setlength{\tabcolsep}{4pt}
\renewcommand{\arraystretch}{1.05}
\caption{Ablation of verification components.}
\label{tab:ablation}
\resizebox{\columnwidth}{!}{%
\begin{tabular}{lccc}
\toprule
Method & ClaimAgr $\uparrow$ & Human-Q/F $\downarrow$
& Resolve $\uparrow$ \\
\midrule
Single-turn only       & 0.384 & 0.612 & 0.562 \\
+\ Multi-agent roles   & 0.422 & 0.590 & 0.585 \\
+\ Role-aware weights  & 0.440 & 0.299 & 0.691 \\
+\ Iterative refinement& \textbf{0.440}
                       & \textbf{0.269}
                       & \textbf{0.703} \\
\bottomrule
\end{tabular}}
\end{table}

Table~\ref{tab:ablation} isolates each component's
contribution. Multi-agent roles improve ClaimAgr
(0.384$\rightarrow$0.422) via complementary observational
scopes. Role-aware weighting delivers the largest gain in
human efficiency (H-Q/F: 0.590$\rightarrow$0.299),
preventing low-confidence evidence from triggering
unnecessary escalations. Iterative refinement provides
additional resolution gains (0.691$\rightarrow$0.703).
The three components thus address distinct failure modes:
complementary scopes reduce agreement failures, role-aware
weights reduce over-escalation, and iteration resolves
claims not settleable in a single pass. Convergence is
rapid, with most gains in the first round; $r_{\max}{=}2$
is sufficient. Dependency-closure re-verification reduces
model calls per accepted edit from 42.3 to 8.7
(4.8$\times$), confirming that correction cost scales with
error scope rather than video length.

\paragraph{User Study}
\label{sec:userstudy}

\begin{figure}[t]
\centering
\includegraphics[width=0.85\linewidth]{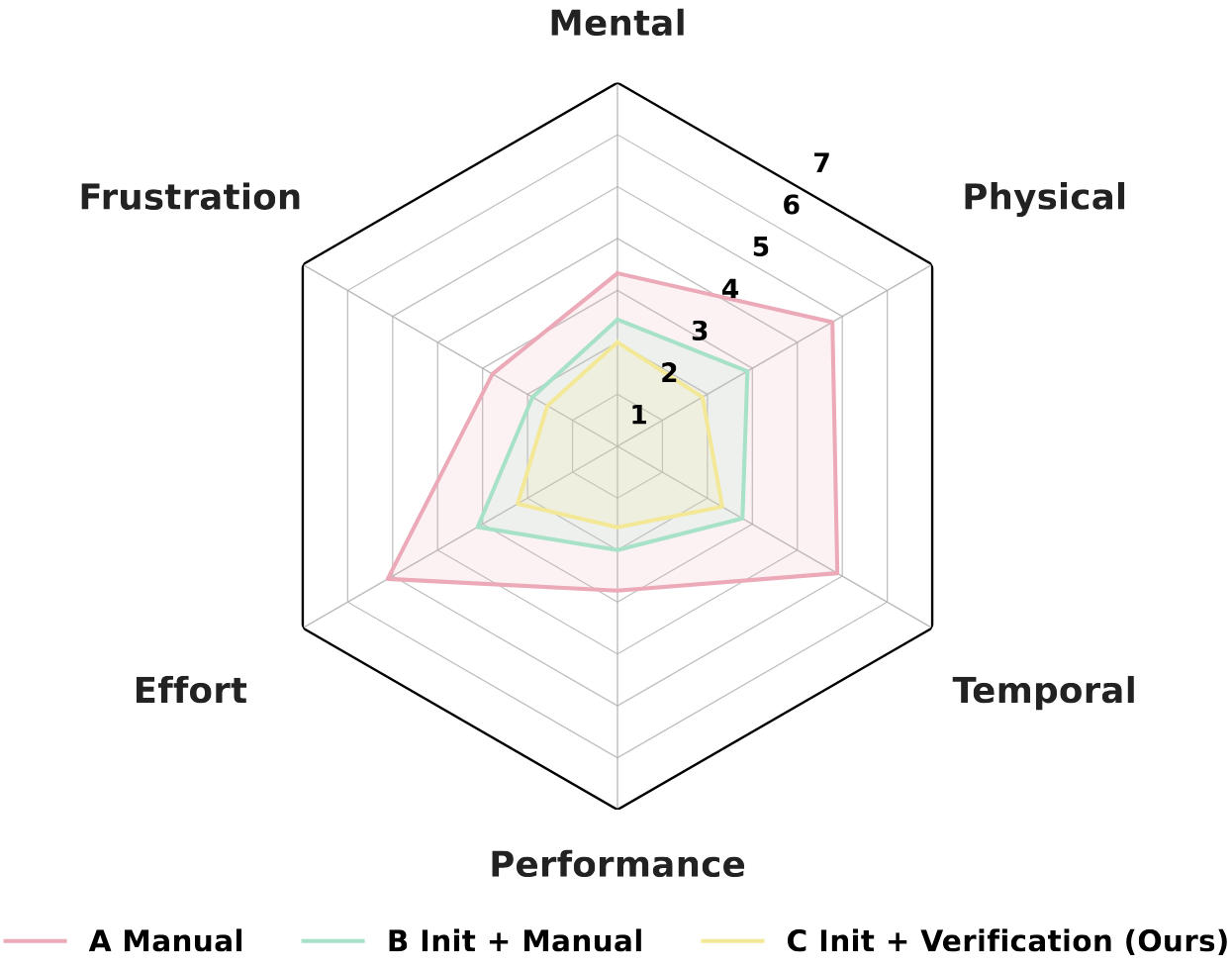}
\caption{NASA-TLX workload comparison (lower is better).}
\label{fig:Figure_TLX}
\end{figure}

We compare three conditions: (A) manual annotation from
raw video, (B) SAM~\cite{kirillov2023segment} initialization
with manual refinement, and (C) SAM initialization with
IMPACT-CYCLE verification. In a pilot study ($n{=}9$),
Condition C achieves the lowest NASA-TLX workload across
all six dimensions (Fig.~\ref{fig:Figure_TLX}). Condition A
imposes significantly higher workload than both B ($p{<}0.05$)
and C ($p{<}0.001$). The B-vs-C gap is directionally
consistent (Cohen's $d{=}0.62$) but does not reach
significance ($p{=}0.096$) at this sample size (estimated
power ${\approx}0.4$--$0.5$); results should be interpreted
as indicative.

\paragraph{Limitations}
VQA evaluation uses the same MLLM for question generation
and assessment, which may inflate reported gains. Human
arbitration in main experiments relies on oracle simulation;
annotator variability at scale remains to be characterized.

%% file: sec/conclusion.tex
\section{Conclusion}

We presented IMPACT-CYCLE, a supervisory multi-agent system
that reframes long-video understanding as iterative
maintenance of a shared, claim-level semantic memory.
Rather than treating error correction as full regeneration,
our framework exposes intermediate reasoning as a
structured, auditable state over which role-specialized
agents verify claims at local, temporal, and global
granularities, and human supervisors intervene only where
automated evidence is insufficient. Experiments on VidOR
demonstrate that verification gains are concentrated in
semantic consistency and downstream reasoning rather than
raw structural overlap, that escalation effort scales
proportionally with graph complexity, and that
dependency-closure re-verification reduces arbitration
cost by 4.8$\times$ relative to full pipeline reruns.
These results suggest that correctability---not generation
quality alone---is the binding constraint in long-video
understanding, and that supervisory verification over
structured semantic memory is a principled path toward
multimodal systems that are not merely accurate but
inspectable, correctable, and trustworthy.